# NGHIÊN CỨU, THIẾT KẾ HỆ THỐNG ĐỊNH VỊ CHO ROBOT DI ĐỘNG TRÊN HỆ ĐIỀU HÀNH ROS

A STUDY AND DESIGN OF LOCALIZATION SYSTEM FOR MOBILE ROBOT BASED ON ROS

**Nguyễn Anh Tú[1], Vũ Công Thành[2], Nguyễn Trọng Hải[3,\*], Nguyễn Trọng Duy[1], Hồ Văn Hoàng[1], Mai Duy Quang[1]**

**TÓM TẮT**

Trong những năm gần đây, robot di động ngày càng được các nhà khoa học quan tâm nghiên cứu bởi những ứng dụng của chúng trong nhiều lĩnh vực khác nhau của đời sống xã hội. Bài báo trình bày giải pháp thiết kế hệ thống định vị cho robot di động ứng dụng hệ điều hành robot (ROS). Hệ thống định vị được thiết kế kết hợp cả phương pháp định vị tương đối và định vị tuyệt đối, trong đó dữ liệu thu được từ các cảm biến encoder, la bàn số và cảm biến quét laser được tổng hợp qua bộ lọc Kalman mở rộng. Hệ thống định vị đã loại bỏ được các sai lệch do các yếu tố môi trường cũng như những sai số tích lũy của phương pháp định vị tương đối. Kết quả thử nghiệm cho thấy độ chính xác và sự ổn định về vị trí và hướng của robot, đáp ứng được các yêu cầu cho các robot làm việc trong nhà.

**Từ khóa:** Robot di động, hệ thống định vị, hệ điều hành robot, bộ lọc Kalman mở rộng.

**ABSTRACT**

In recent years, the mobile robot has been the concern of numerous researcher since they are widely applied in various fields of daily life. This paper applies a virtual robot operating system (ROS) platform to develop a localization system for robot motion. The proposed system is based on the combination of relative and absolute measurement methods, in which the data from the encoder, digital compass, and laser scanner sensor are fused using the extended Kalman filter (EKF). The system also successfully eliminates the errors caused by the environment as well as the error accumulation. The experimental results show good accuracy and stability of position and orientation which can be further applied for the robot working in the indoor environment.

[1]Khoa Cơ khí, Trường Đại học Công nghiệp Hà Nội
[2]Khoa Cơ khí, Trường Đại học Kinh tế - Kỹ thuật Công nghiệp
[3]Viện Cơ khí, Trường Đại học Bách khoa Hà Nội
[\*]Email: hai.nguyentrong@hust.edu.vn

## 1. GIỚI THIỆU

Trong những năm gần đây, robot di động được nghiên cứu và ứng dụng vào nhiều lĩnh vực đa dạng khác nhau như: vận chuyển hàng hóa, cứu hộ, thám hiểm. Sự phát triển của lĩnh vực robot không chỉ dựa trên sự phát triển các giải pháp về phần cứng mà còn phụ thuộc vào khả năng tương thích của phần mềm. Ngày nay, với sự phát triển của các hệ thống thông minh, việc ứng dụng các mã nguồn mở để phát triển hệ thống có ý nghĩa quan trọng, phát huy được sức mạnh của cộng đồng khoa học nhờ đặc tính kế thừa khi cần cấu tái sử dụng lại các thông tin đã tạo trước đó. Do đó, hệ điều hành dành cho robot đang được phát triển và ứng dụng rộng rãi trong những năm gần đây. Hệ điều hành ROS hay hệ điều hành dành cho robot [1, 2], là một giải pháp dành cho robot bao gồm các thư viện và công cụ để điều khiển, có các quy định chung trong việc thiết lập hay giao tiếp giữa các chương trình lẫn nhau. ROS là một hệ điều hành mã nguồn mở được phát triển và cập nhật trong nhiều năm qua [3], nhằm mục đích phát triển công đồng robot, do đó giúp người dùng có thể khai thác những mã nguồn được chia sẻ để phát triển các thuật toán hay tích hợp xây dựng robot. Ngôn ngữ lập trình được sử dụng là các ngôn ngữ phổ biến như C và Python. Trong ROS, hệ thống được tổ chức gồm nhiều node hoạt động cùng nhau và có cơ chế giao tiếp được quy định để truyền nhận dữ liệu. Bên cạnh đó, mỗi node có thể được lập trình với các ngôn ngữ khác nhau, cho phép người dùng có thể sử dụng đa ngôn ngữ phù hợp với hệ thống thiết kế [4].

Các bài toán định vị cho robot là xác định vị trí và hướng của robot so với môi trường làm việc, các thông tin về vị trí phải đủ tin cậy để robot hoạt động chính xác và ổn định [5]. Do đó định vị cho robot đóng vai trò làm tiền đề thực hiện các nhiệm vụ tiếp theo. Các phương pháp định vị cho robot có thể chia làm hai nhóm chính: định vị tương đối và định vị tuyệt đối [6]. Định vị tương đối là xác định vị trí của robot so với vị trí ban đầu tính toán. Phương pháp phổ biến thường được sử dụng trong định vị tương đối là odometry, đây là phương pháp sử encoder đo số vòng quay di chuyển của bánh xe, kết hợp phương trình động học của robot để tính toán vị trí hiện tại so với vị trí ban đầu [7]. Phương pháp này cho phép việc tính toán định vị nhanh, tuy nhiên dao động lớn về vị trí khi hoạt động trong thời gian dài do tích lũy của các lỗi theo thời gian (do độ trượt bánh xe, chu vi bánh). Định vị tuyệt đối là việc xác định vị trí của robot so với với hệ trục tọa độ trong môi trường làm việc. Một trong những phương pháp định vị tuyệt đối là sử dụng các hoa tiêu như các gương phản xạ hay các Barcode gắn cố định trên tường [8]. Trong nghiên cứu của Pierlot đã đề xuất việc sử dụng các góc tương ứng giữa các hoa tiêu, sau đó thiết lập ba đường tròn để tìm ra giao điểm là vị trí hiện tại của robot [9]. Một nghiên cứu khác của Margrit Betke lại sử dụng khoảng cách đến các hoa tiêu để tìm vị trí tương ứng [10]. Điểm chung của các phương pháp sử dụng hoa tiêu này đều yêu cầu việc xác định chính xác vị trí các hoa tiêu trong bộ dữ liệu đã lưu, khi đó sẽ cho độ chính xác cao về vị trí và hướng. Tuy nhiên độ tin cậy và sự ổn định bị ảnh hưởng khi hoạt động tại môi trường có nhiều phản xạ, nơi có độ sáng thay đổi, hay các trường hợp các hoa tiêu tạo nên các hình học đặc biệt như cùng nằm trên đường tròn, đường thẳng. Ngoài ra việc can thiệp vào môi trường hoạt động gây mất tính linh hoạt trong quá trình điều hướng cho robot. Chính vì vậy, phương pháp định vị sử dụng bản đồ số đang được sử dụng nhiều hiện nay [11] do việc định vị dựa trên thông tin của môi trường làm việc nên tạo ra được tính linh hoạt trong quá trình di chuyển. Ngoài ra còn cung cấp được thông tin về các vật cản trong quá trình di chuyển tạo cơ sở giải quyết bài toán tránh vật cản hay thiết kế quỹ đạo sau này. Tuy nhiên, do việc xác định vị trí và hướng dựa trên các phép so sánh với bản đồ đã tạo trước, vì vậy thời gian đáp ứng phụ thuộc vào tốc độ tính toán của hệ thống. Khi môi trường có sự thay đổi lớn, hay tốc độ tính toán không đáp ứng được thường gây ra các hiện tượng lệch thông tin vị trí và hướng của robot [12]. Có thể thấy, các nghiên cứu trên chủ yếu được xây dựng dựa trên một cảm biến, do đó các thông tin định vị dễ bị ảnh hưởng khi làm việc tại nhiều môi trường khác nhau, dẫn đến các sai số không mong muốn.

Trong nghiên cứu này, bài báo trình bày việc xây dựng hệ thống định vị cho robot di động dựa trên hệ điều hành dành cho robot (ROS). Giải pháp thiết kế kết hợp cả phương pháp định vị tương đối và định vị tuyệt đối, dữ liệu từ các biến được tổng hợp bằng bộ lọc Kalman mở rộng. Các kết quả được thử nghiệm trên mô hình thực tế, cho thấy hệ thống đạt được sự ổn định, độ chính xác về vị trí và hướng khi hoạt động trong các môi trường làm việc trong nhà.

## 2. XÂY DỰNG HỆ THỐNG ĐỊNH VỊ TRÊN HỆ ĐIỀU HÀNH ROS

Để điều khiển robot chuyển động theo quỹ đạo mong muốn và đi đến điểm đích, tại mỗi thời điểm cần xác định được thông tin về vị trí và hướng (góc) của robot. Hệ thống định vị được thiết kế trong bài báo kết hợp thông tin từ các cảm biến encoder, la bàn số và cảm biến quét laser. Sơ đồ cấu trúc của hệ thống trên ROS được phát triển dựa trên các node, mỗi node được tạo ra có nhiệm vụ và chức năng khác nhau. Các thông tin trao đổi giữa từng node được giao tiếp qua các tin nhắn, các thông tin truyền nhận theo các giao thức quy định trước và có chung một chuẩn phục vụ

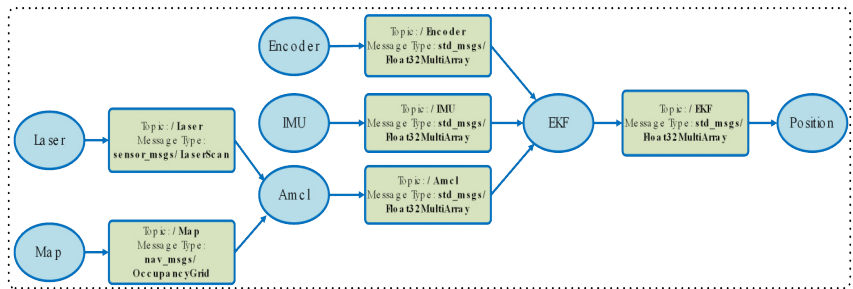

Hình 1. Sơ đồ tổ chức các node trong ROS

việc giao tiếp với các gói dữ liệu bên ngoài. Bên cạnh đó, công cụ hỗ trợ việc kiểm tra các giải thuật và theo dõi dữ liệu có tính trực quan và đảm bảo dữ liệu được xử lý theo thời gian thực. Sơ đồ cấu trúc hệ thống được thiết kế trên ROS gồm các node và quá trình truyền nhận dữ liệu được mô tả như trong hình 1.

Trong hệ thống được thiết kế, node Encoder thực hiện việc tính toán vận tốc dài và vận tốc góc của robot dựa trên sự thay đổi số xung của các encoder. Node IMU có chức năng đọc dữ liệu của la bàn số thông qua giao thức UART và trả về dữ liệu đầu ra là góc và vận tốc góc của robot. Thông tin của bản đồ cục bộ của robot được xác định qua các đám mây điểm từ cảm biến Laser, dữ liệu này được node Amcl đối sánh với bản đồ của môi trường làm việc được nhập trước vào hệ thống để tìm ra các xác suất vị trí của robot trên bản đồ thực. Các thông tin định vị từ ba loại cảm biến trên được đưa vào vào node EKF để xác định vị trí và hướng của robot tại mỗi thời điểm để phục vụ quá trình điều khiển chuyển động của robot. Trong hệ thống xây dựng trên ROS, mỗi node có nhiệm vụ và chức năng độc lập, do đó khi cần thay đổi hoặc điều chỉnh, chỉ cần thực hiện trên từng node mà không cần xây dựng lại toàn bộ chương trình. Chức năng của các node được mô tả chi tiết trong bảng 1.

Bảng 1. Chức năng từng node của hệ thống ROS

| STT | Tên node | Chức năng |
|---|---|---|
| 1 | Encoder | Chuyển đổi số xung của hai bánh về vận tốc dài và vận tốc góc của robot trên cơ sở phương trình động học thuận. |
| 2 | IMU | Đọc tín hiệu từ la bàn số để tính toán vận tốc dài và vận tốc góc của robot. |
| 3 | Laser | Đọc dữ liệu đám mây điểm của cảm biến laser đưa ra bản đồ cục bộ. |
| 4 | Map | Cung cấp dữ liệu của bản đồ môi trường được cập nhật trước. |
| 5 | Amcl | Sử dụng thuật toán Acml để đối sánh thông tin giữa đám mây điểm hiện tại và bản đồ làm việc toàn cục để đưa ra được vị trí và hướng của robot. |
| 6 | EKF | Kết hợp các tín hiệu định vị từ encoder, la bàn số, bản đồ số để ước lượng về vị trí và góc của robot. |
| 7 | Position | Chứa thông tin về vị trí và góc của robot phục vụ các nhiệm vụ công nghệ khác như theo dõi vị trí robot, lấy dữ liệu phục vụ quá trình điều khiển chuyển động. |

## 3. THUẬT TOÁN KALMAN MỞ RỘNG

Do đặc tính phi tuyến của mô hình động học robot và đặc tính làm việc riêng biệt của từng cảm biến, nên bộ lọc Kalman mở rộng được sử dụng (EKF) để tổng hợp thông tin định vị từ các cảm biến. Mô hình chuyển đổi trạng thái và đo lường được biểu diễn dưới dạng:

$$X_k = f(X_{k-1}) + W_{k-1} \quad \text{và} \quad z_k = h(X_k) + V_k \quad (1)$$

Trong đó, f và h lần lượt là hàm phi tuyến mô tả trạng thái của hệ thống và mô hình đo lường. Nhiễu quá trình $W_{k-1}$ và nhiễu đo lường $V_k$ là được giả thuyết phân bố theo quy luật Gaussian với giá trình trung bình bằng 0 với ma trận hiệp phương sai: $W_k \sim N(0, Q_k)$ và $V_k \sim N(0, R_k)$.

Ma trận Jacobian tuyến hình hoá f và h quanh điểm làm việc được xác định:

$$F_{k-1} = \frac{\partial f(X)}{\partial X}\bigg|_{\hat{X}_{k-1}^-} \quad (2)$$

$$F_k = \begin{bmatrix} 1 & 0 & -v_k \Delta t \sin(\theta_R)_k & \Delta t \cos(\theta_R)_k & 0 \\ 0 & 1 & v_k \Delta t \cos(\theta_R)_k & \Delta t \sin(\theta_R)_k & 0 \\ 0 & 0 & 1 & 0 & \Delta t \\ 0 & 0 & 0 & 1 & 0 \\ 0 & 0 & 0 & 0 & 1 \end{bmatrix} \quad (3)$$

$$H_k = \frac{\partial h(X)}{\partial X}\bigg|_{\hat{X}_k^-} \quad (4)$$

Bản đồ số trả về các thông tin về vị trí và hướng của robot trong hệ tọa độ toàn cục, ma trận đo lường được biểu diễn dưới dạng:

$$h_{map} = \begin{bmatrix} x_{map} \\ y_{map} \\ \theta_{map} \end{bmatrix} = \begin{bmatrix} (x_R)_k \\ (y_R)_k \\ (\theta_R)_k \end{bmatrix} \text{ và } H_{map} = \begin{bmatrix} 1 & 0 & 0 & 0 & 0 \\ 0 & 1 & 0 & 0 & 0 \\ 0 & 0 & 1 & 0 & 0 \end{bmatrix} \quad (5)$$

$$z_k = H_{map} X_k + V_k \quad (6)$$

Thông tin về vận tốc dài v và vận tốc góc ω của robot được xác định từ cảm biến encoder với ma trận đo lường được viết dưới dạng:

$$h_{encoder} = \begin{bmatrix} v_{encoder} \\ \omega_{encoder} \end{bmatrix} = \begin{bmatrix} v_k \\ \omega_k \end{bmatrix} \text{ và } H_{encoder} = \begin{bmatrix} 0 & 0 & 0 & 1 & 0 \\ 0 & 0 & 0 & 0 & 1 \end{bmatrix} \quad (7)$$

$$z_k = H_{encoder} X_k + V_k \quad (8)$$

Ma trận đo lường thông tin về góc $\theta_R$ và vận tốc góc ω được mô tả dưới dạng:

$$h_{compass} = \begin{bmatrix} \theta_{compass} \\ \omega_{compass} \end{bmatrix} = \begin{bmatrix} (\theta_R)_k \\ \omega_k \end{bmatrix}$$

$$\text{và } H_{compass} = \begin{bmatrix} 0 & 0 & 1 & 0 & 0 \\ 0 & 0 & 0 & 0 & 1 \end{bmatrix} \quad (9)$$

$$z_k = H_{compass} X_k + V_k \quad (10)$$

Quá trình tính toán được thực hiện gồm các bước dự đoán và cập nhật như sau:

Dự đoán:

$$\hat{X}_k^- = f(\hat{X}_{k-1}^+, u_k)$$

$$P_k^- = F_k P_{k-1}^+ F_k^T + Q_k$$

Cập nhật:

$$K_k = \frac{P_k^- H_k^T}{H_k P_k^- H_k^T + R_k}$$

$$\hat{X}_k^+ = \hat{X}_k^- + K_k \left( z_k - h(\hat{X}_k^-) \right)$$

$$P_k^+ = P_k^- - K_k H_k P_k^-$$

## 4. THIẾT KẾ PHẦN CỨNG

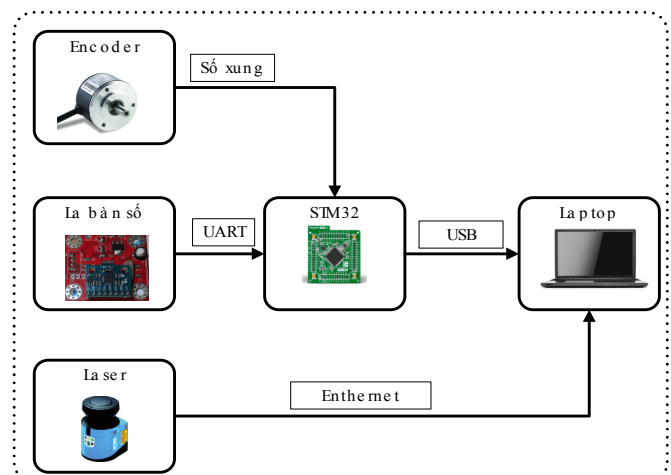

Hình 2. Sơ đồ mô hình phần cứng của robot

Mô hình phần cứng hệ thống định vị được mô tả như hình 2, gồm bộ điều khiển trung tâm được sử dụng là một laptop có cài hệ điều hành ROS với mã phiên bản Kinetic. Hệ thống cảm biến được sử dụng trên robot bao gồm cảm biến quét laser, la bàn số và encoder. Cảm biến laser thu thập thông tin từ môi trường và gửi dữ liệu đám mây điểm lên laptop thông qua giao thức truyền thông enthernet. Vi điều khiển STM32 đóng vai trò thu nhận và xử lý thông tin từ cảm biến encoder và la bàn số để gửi về laptop thông qua giao tiếp USB. Đặc tính kỹ thuật của các hiết bị được mô tả như trong bảng 2 và sơ đồ kết nối hệ thống được mô tả như trong hình 3.

Bảng 2. Các thiết bị của hệ thống

| STT | Tên thiết bị | Đặc tính kỹ thuật |
|---|---|---|
| 1 | Cảm biến laser | - Khoảng cách đo: 0,5m - 50m<br>- Góc quét: 2700<br>- Sai số đo dài: ±8mm (khoảng cách đo nhỏ hơn 30m); ±10mm (khoảng cách đo từ 30 đến 50m)<br>- Sai số đo góc: ±0,040 (khoảng cách đo nhỏ hơn 30 m); ±0,060 (khoảng cách đo từ 2 đến 50m)<br>- Tần số làm việc: 25Hz ± 5%<br>- Truyền thông Enthernet: 100 MBit/s, TCP/IP<br>- Điện áp làm việc: 10,8V đến 30VDC |

| 2 | Encoder | - Encoder Omron tương đối E6B2-CWZ6C 500P/R<br>- Điện áp hoạt động: 5 đến 24VDC<br>- Độ phân giải: 500 xung/vòng<br>- Tần số đáp ứng: 100kHz |
|---|---|---|
| 3 | La bàn số | - Trả về thông tin góc quay của đối tượng với độ phân giải 0.1$^0$<br>- Tín hiệu trả về là một số nguyên 16bit thông qua giao thức UART với tốc độ baud là 115200bps<br>- Trên module có nút nhấn để có thể hiệu chỉnh lại |
| 4 | Vi điều khiển | - Tần số hoạt động: 168MHz<br>- Số chân tín hiệu: 100 chân<br>- Điện áp hoạt động: 3,3V<br>- Ngoại vi: Timer, SPI, USART, I2C… |
| 5 | Máy tính | - Bộ xử lý: Intel Core i5-6200 U<br>- Ram: 4GB<br>- Hệ điều hành sử dụng: Ubuntu 16.04<br>- Phiên bản ROS cài đặt: Kinetic |

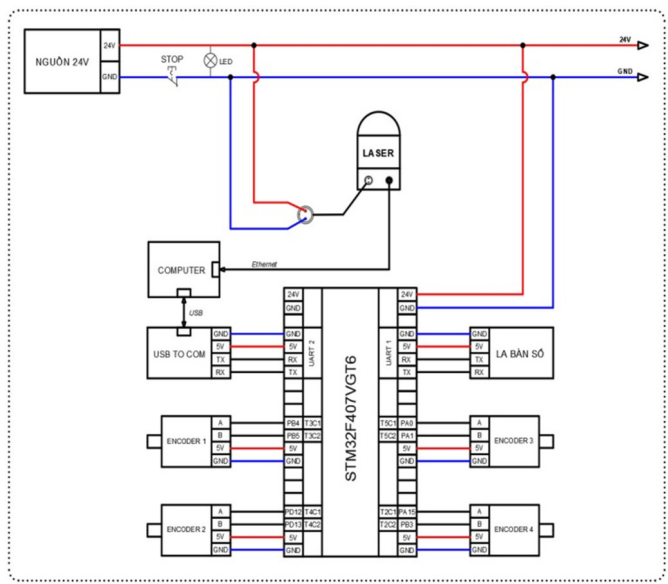

Hình 3. Sơ đồ kết nối hệ thống.

## 5. THỬ NGHIỆM VÀ ĐÁNH GIÁ HỆ THỐNG

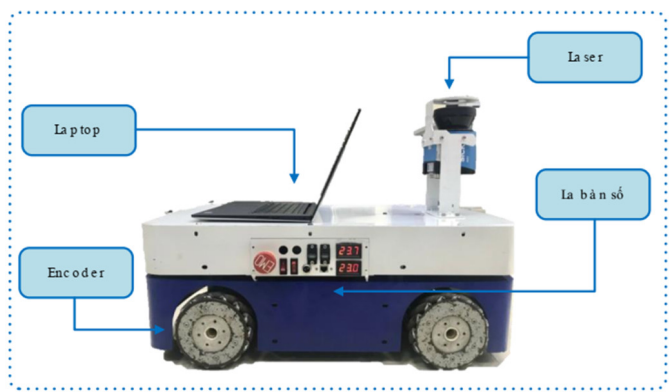

Hình 4. Mô hình robot thử nghiệm

Để đánh giá độ chính xác và hoạt động của giải pháp đề xuất, hệ thống định vị được tích hợp lên robot di động có cấu trúc dạng bánh đa hướng mecanum như hình 4. Quỹ đạo thử nghiệm có dạng hình chữ nhật với quãng đường chuyển động là 78m. Để xây dựng bản đồ môi trường làm việc, robot được điều khiển ở chế độ bằng tay di chuyển dọc theo quỹ đạo thiết kế để quét hình ảnh môi trường làm việc thông qua cảm biến laser (hình 5), bản đồ sau đó được cập vào hệ thống định vị trước khi cho robot chạy tự động. Robot được thử nghiệm chuyển động với vận tốc 0,8m/s, các tín hiệu định vị được tính toán ở tần số 17Hz và được cập nhật lên hệ thống điều khiển của robot. Đáp ứng vị trí và góc di chuyển được mô tả lần lượt tại hình 6 và 7 với sai số trung bình về vị trí và hướng tính theo phương pháp Root Mean Square Error (công thức 11, 12) lần lượt là 0,0203m và 0,6939$^0$.

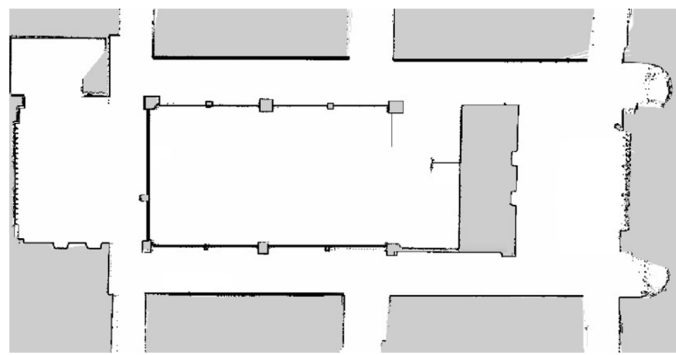

Hình 5. Bản đồ toàn cục của môi trường thử nghiệm

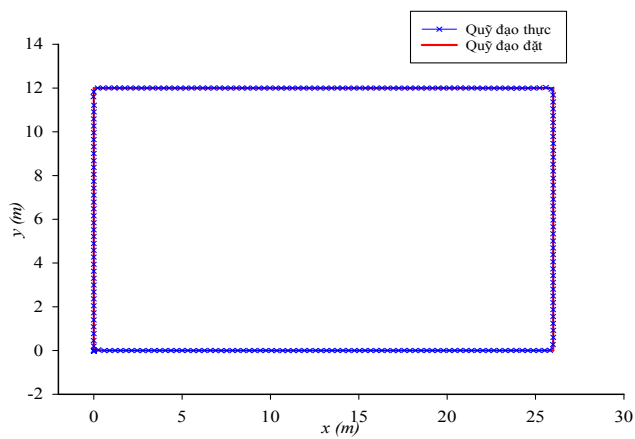

Hình 6. Đáp ứng vị trí của robot

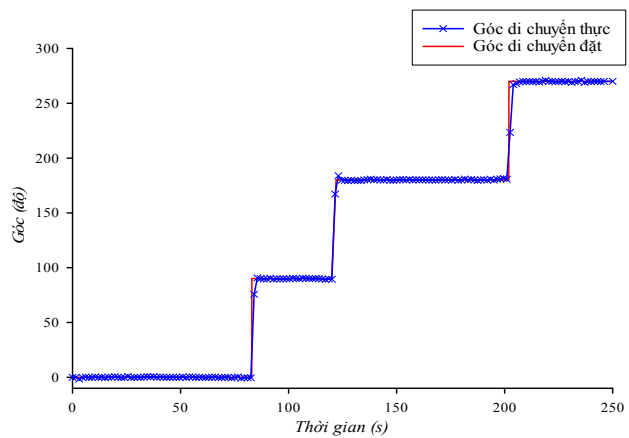

Hình 7. Đáp ứng góc của robot

$$e_{dist} = \sqrt{\frac{1}{n}\sum_{k=1}^{n}(x_{ref}-x_{sensor})^2 + (y_{ref}-y_{sensor})^2} \quad (11)$$

$$e_{\theta} = \sqrt{\frac{1}{n}\sum_{k=1}^{n}(\theta_{ref}-\theta_{sensor})^2} \quad (12)$$

Các sai lệch vị trí theo hai phương x, y được mô tả lần lượt trên hình 8 và 9 với sai lệch lớn nhất lần lượt là 0,03m và 0,028m. Kết quả thử nghiệm cho thấy, khi robot chuyển động trên các đoạn theo phương x hoặc y thì các sai lệch theo phương dịch chuyển dao động quanh giá trị 0,02m, trong khi đó sai lệch với phương vuông góc với phương dịch chuyển dao động quanh giá trị 0,0016m. Điều này cho thấy robot di chuyển tương đối ổn định, không bị hiện tượng rung lắc, các sai lệch gây ra bởi các nguyên nhân chính như quán tính xe, hiện tượng trượt giữa bánh xe và môi trường làm việc. Bên cạnh đó, hình 10 cho thấy các sai lệch góc (hướng của robot) có hiện tượng dao động mạnh tại các vị trí có sự thay đổi về hướng dịch chuyển là các vị trí góc của quỹ đạo hình chữ nhật với sai số góc lớn nhất là 3,3$^0$.

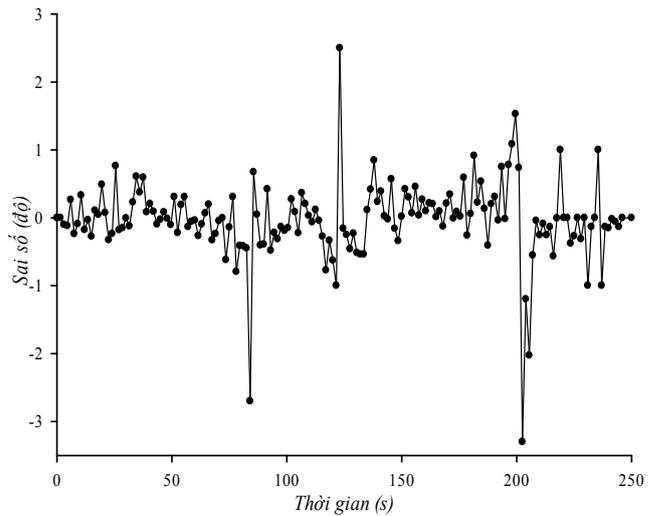

Hình 10. Sai số góc di chuyển

Từ kết quả thí nghiệm cho thấy, mặc dù trong quá trình thử nghiệm vẫn xuất hiện các sai lệch về vị trí của robot, tuy nhiên các sai lệch có giá trị thay đổi lớn chỉ xuất hiện tại một số thời điểm robot thực hiện thay đổi đột ngột về hướng chuyển động. Khi robot di chuyển với hướng ổn định thì các sai lệch chủ yếu phát sinh theo phương chuyển động, đồng nghĩa với giải pháp điều khiển vẫn chưa đáp ứng hoàn toàn với các yếu tố như quán tính, sai lệch trong lắp ráp, chế tạo nhưng đáp ứng tốt về phương. Tuy nhiên hệ thống định vị đã loại bỏ được các yếu tố sai lệch tích lũy, các yếu tố ảnh hưởng của môi trường và cho độ chính xác đáp ứng được yêu cầu cơ bản cho hoạt động của robot di động làm việc ở môi trường trong nhà.

## 6. KẾT LUẬN

Bài báo đã trình bày một giải pháp thiết kế hệ thống định vị cho robot di động dựa trên hệ điều hành robot ROS, bao gồm cả phần cứng và phần mềm. Hệ thống định vị kết hợp cả giải pháp định vị tương đối và định vị tuyệt đối giúp nâng cao độ chính xác, tính ổn định khi môi trường làm việc có các yếu tố gây nhiễu và loại bỏ sai số tích lũy của các loại cảm biến tương đối. Hệ thống thử nghiệm được tích hợp lên một robot di động dạng đa hướng và thử nghiệm hoạt động ở môi trường làm việc trong nhà. Kết quả thử nghiệm cho thấy hệ thống hoạt động ổn định, sai số về vị trí và hướng lần lượt 0,0203m và 0,6939$^0$ đáp ứng được cho các robot làm việc trong nhà. Bên cạnh đó, giải pháp thiết kế hệ thống trên ROS cho phép người dùng có thể linh hoạt trong việc điều chỉnh, cấu trúc lại theo từng ứng dụng cụ thể.

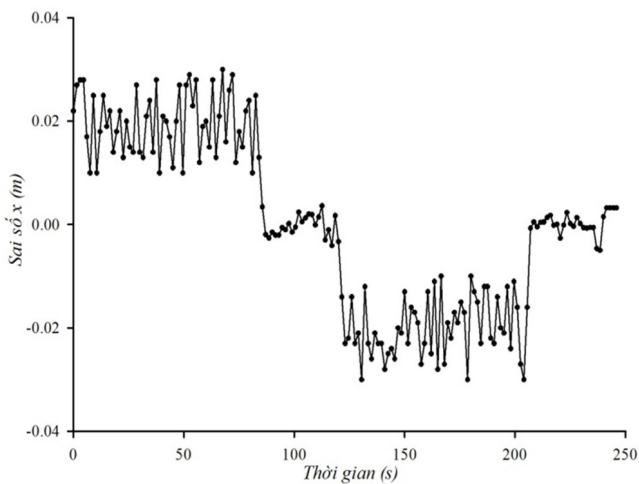

Hình 8. Sai số vị trí theo phương x

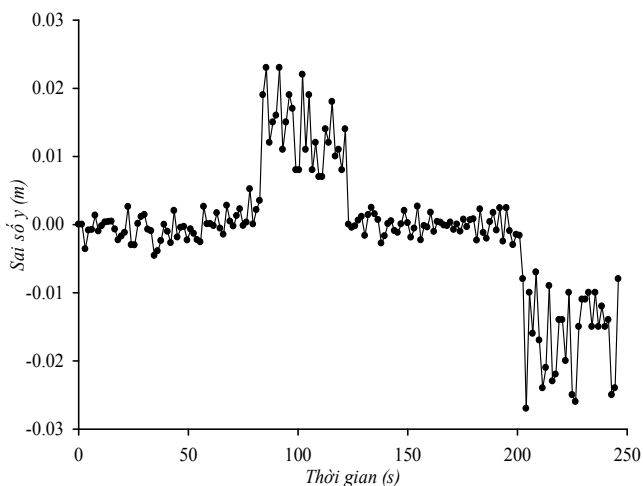

Hình 9. Sai số vị trí theo phương y

**AUTHORS INFORMATION**

**Nguyen Anh Tu[1], Vu Cong Thanh[2], Nguyen Trong Hai[3], Nguyen Trong Duy[1], Ho Van Hoang[1], Mai Duy Quang[1]**

[1]Faculty of Mechanical Engineering, Hanoi University of Industry

[2]Faculty of Mechanical Engineering, University of Economics - Technology for Industries

[3]School of Mechanical Engineering, Hanoi University of Science and Technology